\pdfoutput=1
\documentclass[11pt]{article}

\usepackage{EACL2023}

\usepackage{times}
\usepackage{latexsym}

\usepackage[T1]{fontenc}
\usepackage[utf8]{inputenc}

\usepackage{microtype}
\usepackage{inconsolata}
\usepackage[frozencache,cachedir=.]{minted}

\usepackage{xurl}
\usepackage{times}
\usepackage{latexsym}
\usepackage{enumitem}
\usepackage{paralist}
\usepackage{caption}
\usepackage{subcaption}

\usepackage{comment}
\usepackage[flushleft]{threeparttable}
\usepackage{graphicx}
\usepackage{amsmath}
\usepackage{tabularx}
\usepackage{mathrsfs}
\usepackage{multirow, makecell}
\usepackage{algorithm}
\usepackage{enumitem}
\usepackage{xcolor}
\usepackage{listings}
\usepackage{hyperref}
\usepackage{arydshln}
\usepackage{mathtools}

\usepackage{amssymb}% http://ctan.org/pkg/amssymb
\usepackage{pifont}% http://ctan.org/pkg/pifont
\newcommand{\cmark}{\ding{51}}%
\newcommand{\xmark}{\ding{55}}%

\usepackage[utf8]{inputenc}
\usepackage{cleveref}
\usepackage{minted}
\crefname{section}{§}{§§}
\Crefname{section}{§}{§§}

\usepackage{xcolor} 
\usepackage{tcolorbox}

\definecolor{blueish}{RGB}{250, 250, 255}
\definecolor{greenish}{RGB}{250, 255, 250}
\newtcbox{\inlinebox}[1][]{
 box align=base,
 nobeforeafter,
 colback=blueish,
 size=small,
 left=0pt,
 right=0pt,
 boxsep=2pt,
 #1}

\definecolor{MyColor}{RGB}{50, 100, 250}
\definecolor{Orange}{RGB}{244, 101, 66}
\definecolor{Red}{RGB}{255, 0, 0}
\definecolor{Green}{RGB}{0, 255, 0}
\definecolor{Blue}{RGB}{0, 0, 255}

\newcommand{\tool}{ERA~}
\newcommand{\toolnospace}{ERA}
\newcommand{\toolextnospace}{ERA-D}
\newcommand{\toolext}{ERA-D~}

\DeclareMathOperator*{\argmax}{arg\,max}

\renewenvironment{itemize}[1]{\begin{compactitem}#1}{\end{compactitem}}

\usepackage{arydshln}
\usepackage[belowskip=0pt,aboveskip=5pt]{caption}
\usepackage{fdsymbol}

\usepackage{titlesec}
\titlespacing{\paragraph}{%
  0pt}{%              left margin
  0.2\baselineskip}{% space before (vertical)
  1em}%

\title{Retrieval Enhanced Data Augmentation for \\ Question Answering on Privacy Policies}

\author{
Md Rizwan Parvez$^\S$, Jianfeng Chi$^\dagger$, Wasi Uddin Ahmad$^\S$,  \\
\textbf{Yuan Tian$^\S$, Kai-Wei Chang$^\S$} \\
$^\S$University of California, Los Angeles,
$^\dagger$University of Virginia\\
$^\S$\texttt{{\{rizwan,wasiahmad,yuant,kwchang\}@ucla.edu}},
$^\dagger$\texttt{{jc6ub@virgina.edu}}
}

\begin{document}
\maketitle
\begin{abstract}
% Question answering (QA) systems for privacy policies help users in finding the relevant information from long and verbose policy documents.  
Prior studies in privacy policies frame the
question answering (QA) task as identifying the most relevant text segment or a list of sentences from a policy document given a user query. 
% However, annotating such a dataset is challenging as it requires specific domain expertise (e.g., law academics). 
% Even if we manage a small-scale one, a bottleneck that remains is that the labeled data are heavily imbalanced (only a few segments are relevant) -- limiting the gain in this domain. 
Existing labeled datasets are heavily imbalanced (only a few relevant segments), limiting the QA performance in this domain. 
% Under-sampling the abundant (irrelevant) examples make the data even more insufficient.
% to build statistical NLP models.
% \notewa{What if the readers ask, why don't we only consider equal number of positives and negatives? Let's say, for a particular question, there are 5 answer sentences. Then we can randomly pick 5 non-answer sentences. Then there is no data imbalance issue, right? How should we tackle this question?}
% Therefore, 
In this paper, we develop a data augmentation framework based on ensembling retriever models that captures the relevant text segments from unlabeled policy documents and expand the positive examples in the training set. 
% Uniquely, our retrievers are built upon multiple different pre-trained language models (LM) and cascaded with noise reduction oracles. 
In addition, to improve the diversity and quality of the augmented data, we leverage multiple pre-trained language models (LMs) and cascade them with noise reduction filter models.
Using our augmented data on the PrivacyQA benchmark, we elevate the existing baseline by a large margin (10\% F1) and achieve a new state-of-the-art F1 score of 50\%. Our ablation studies provide further insights into the effectiveness of our approach.
% \notewa{We should start by talking about how QA task is framed in PrivacyQA (fundamentally it is an answer sentence selection task). Then talk about the label imbalance issue. Finally talk about our proposal.}

\end{abstract}

\section{Introduction}
\label{sec:intro}

% in this domain long and complex documents difficult for users to read
% and understand,  A question answering (QA) system can
% . Prior studies
% in this domain frame the QA task as retrieving
% the most relevant text segment or a list of sentences from the policy document given a question. 
% Question answering (QA) systems for privacy policies help users in finding the relevant information from long and verbose policy documents.
% To prevent  \notewa{seems like this is the only reason why the users should read/understand policies}, 

% to find if conditions outlined in the privacy policy acceptable.

% Understanding privacy policies of is crucial for users to find the conditions outlined in its privacy policy acceptable.
% users as it empowers them to learn about the information that matters to them

% decide to use a company’s products or services only if they find the conditions outlined in its privacy policy acceptable.

% Understanding privacy policies that describe how user data is collected, managed, and used by the respective service providers is crucial for determining if the conditions outlined are acceptable. 
Privacy policies describe how service providers collect, manage, and use their users' data. Understanding them is crucial for users as they can determine if the conditions outlined are acceptable. 
Policy documents, however, are  lengthy, verbose, equivocal, and hard to understand~\cite{mcdonald2008cost, reidenberg2016ambiguity}. Consequently, they are often ignored and skipped by users~\cite{federal2012protecting, gluck2016short}. 

\begin{table}[t]
\centering
% \vspace{4mm}
\resizebox{\linewidth}{!}{%
%  \small
% \footnotesize
% \sciptsize
\def\arraystretch{1.1}%
\begin{tabular}{p{\linewidth}}
\hline
Segmented policy document $\textbf{\em S}$ \\
\hline
($\textbf{\em s}_1$) {\color{red} We do not sell or rent your personal information to third parties for their direct marketing purposes without your explicit consent.}
% ... 
($\textbf{\em s}_n$) ...We will not let any other person, including sellers and buyers, contact you, other than through your ... \\
% to others except as disclosed in this Policy or ... \\
% \medskip
\hline
% \medskip
Queries $\textbf{\em I}$ annotating the {\color{red} red} segment as {\color{blue} irrelevant} \\
\hline
($\textbf{\em i}_1$) How does Fiverr protect freelancers' personal information?
($\textbf{\em i}_2$) What type of identifiable information is passed between users on the platform? \\
% c) Who can see my information? 
% \medskip
\hline
% \medskip
Queries $\textbf{\em R}$ annotating the {\color{red} red} segment as {\color{blue} relevant} \\
\hline
($\textbf{\em r}_1$) What are the app's permissions? ($\textbf{\em r}_2$) What type of permissions does the app require? \\
% \makeskip
\hline
% \medskip
Queries $\textbf{\em D}$ that annotators {\color{blue} disagree} about relevance\\
\hline
% \medskip
($\textbf{\em d}_1$) Do you sell my information to third parties? ($\textbf{\em d}_2$)
Is my information sold to any third parties? \\
% \makeskip
\hline
\end{tabular}
}
% \vspace{-1mm}
\caption{
% \small
QA (sentence selection) from a policy document $\textbf{\em S}$. {\bf Sensitive}: For queries $\textbf{\em R}$ and $\textbf{\em I}$, annotators at large tagged sentence $\textbf{\em s}_1$ as relevant, and irrelevant respectively. On the other hand, sentence $\textbf{\em s}_n$, though analogous to $\textbf{\em s}_1$ in meaning, was never tagged as relevant. {\bf Ambiguous}: For queries $\textbf{\em D}$, experts interpret $\textbf{\em s}_1$ differently and disagree on their annotations.
}
\label{table:qa-task}
% \vspace{-6mm}
\end{table}

Building question answering (QA) systems for privacy policies is a stepping stone to allow users to ask questions about their rights.
Prior works \cite{harkous2018polisis, Ravichander2019Question} framed the QA task as a sentence selection task, essentially a binary classification task that identifies if a policy text segment is relevant to a question.
% To help the users better understand their rights, privacy policy QA is framed to answer sentence selection task,
% % from the long and verbose policies
% essentially a binary classification task to identify if a policy text segment is relevant or not~\cite{harkous2018polisis}.
% \notewa{clarify it saying the task is answer sentence selection.}
% However, annotating policy documents requires expertise and domain knowledge, and hence, it is costly and hard to obtain.
% For the only existing one\footnote{\citet{ahmad2020policyqa} is another one with a different task.
% See Sec.~\ref{sec:related-work} for more detail.
% }, PrivacyQA dataset~\cite{Ravichander2019Question},
% Even if we are fortunate enough to annotate a small-size corpus,
% a bottleneck that prevails is that it is heavily imbalanced (only a few are positive), which limits the gain. 
Since policy documents consist of many sentences and typically a few are relevant to a question,\footnote{PrivacyQA~\cite{Ravichander2019Question}
dataset has 1,350 questions with an average number of answer sentences is 5, while the average length of policy documents is 138 sentences.} the classification data is imbalanced.
% Moreover, as most texts in policy documents are not relevant, the data is heavily imbalanced.
% For example, the only existing dataset, PrivacyQA~\cite{Ravichander2019Question}
% % \footnote{\citet{ahmad2020policyqa} is another one with a different setting.
% % % See Sec.~\ref{sec:related-work} for more detail.
% % }
% has 1,350 questions in the training dataset, and the average number of answer sentences is 5, while the average length of policy documents is 138 sentences. 
% Its authors train an answer sentence selection model 
% (which is a binary classification model in a nutshell)
% using 7K positive with 178K negative sentences that underperforms human performance by 0.3 F1. Sub-sampling the negative examples makes the data even more insufficient to learn from (e.g., using the equal number of positive and negative examples (7k), the F1 score drops by 10\%).  \notesc{10\% or 0.1 F1?}
% \notesc{We can add another bottleneck here (disagreement) between annotators.}
In this work, we attempt to mitigate the data imbalance by augmenting positive QA examples.
Specifically, we develop automatic retrieval models to supplement relevant policy sentences for each user query. 
We keep the queries unchanged as they are usually limited to a few forms only~\cite{wilson2016creation}.
% usually have minor variants and are.

Unlike other domains, augmenting privacy policy statements is very challenging. 
First, they often describe similar information~\citep{hosseini2016lexical}. Thus, their annotations are sensitive to small changes in the text (see Table \ref{table:qa-task}) which may not be tackled using the existing augmentation methods  based on data synthesis (e.g., mixup~\cite{zhang2018mixup}, back-translation~\cite{edunov-etal-2018-understanding}). For example, \citet{kumar2020data} identifies that even linguistically coherent instances augmented via generative models do not preserve the class labels well.\footnote{To verify, in our preliminary study, for each positive example in the PrivacyQA training set, we augment a new synthesized positive example by en-zh-en back-translation using Google Translator API, and the performance drops by 3\% F1.}
Hence, to reflect the data properties, we consider a retrieval-based approach to augment the raw policy statements. Given a pre-trained LM and a small QA dataset, we first build a dense sentence retriever~\cite{karpukhin2020dense}. 
% parvez-etal-2021-retrieval-augmented}.
% and the available training data.
Next, leveraging an unlabeled policy corpus with 0.6M sentences crawled from web applications, we perform a coarse one-shot sentence retrieval for each query in the QA training set.
% preserve the relevant retrievals 
To filter the noisy candidates retrieved,\footnote{We refer the misclassified candidates as noisy retrievals.} we then train a QA model (as a filter model) using the same pre-trained LM and data and couple it with the retriever. 

\begin{figure}[t]
\centering
% \hspace{5pt}
\includegraphics[width=0.48\textwidth]{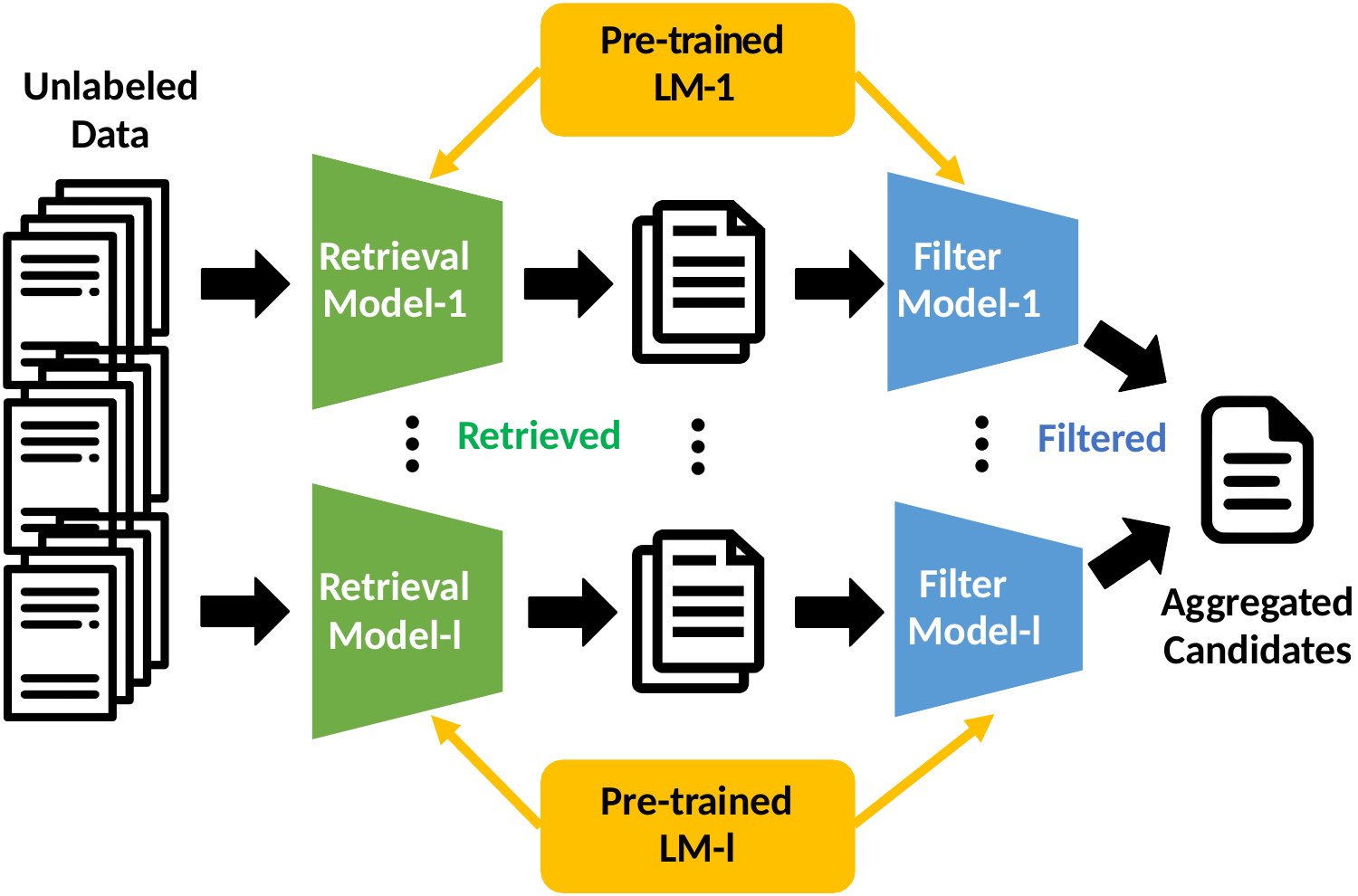}
\caption{
Our framework. Given a pre-trained LM, we train (i) a retriever, (ii) a QA model (filter) both on the small-size labeled data. From an unlabeled corpus, we first, retrieve the coarse relevant sentences (positive examples) for the queries in the training set and use the filter model to filter out noisy ones. We repeat this for multiple different pre-trained LMs. Finally, we aggregate them to expand the positive examples in the training set and learn any user-defined final QA model.
% \vspace{-0.3cm}
}
\label{fig:model}
% \vspace{-0.2cm}
\end{figure}

Second, privacy policies are ambiguous; even skilled annotators dispute their diverse interpretations, e.g., for at least 26\% questions in PrivacyQA, experts disagree on their annotations (see Table \ref{table:qa-task}). Therefore, a single retriever model may not capture all sorts of relevant policy segments written in various diversified ways. To combat this insufficient data diversity, we propose a novel retriever ensemble technique. Different pre-trained models learn distinct language representations due to their pre-training objectives, and hence, retriever models built on them can retrieve a disjoint set of candidates (verified in Section~\ref{sec:exp}). 
% For example, using three different retrievers, only 15 candidates out of around 7K  overlap (details in Section ).
Therefore, we build our retrievers and filter models based on multiple different pre-trained LMs (See Figure~\ref{fig:model}).
% followed by augmenting them with the annotated QA training set (See Figure~\ref{fig:model}). 
Finally, we train a user-defined QA model on the aggregated corpus using them. 

% We evaluate our method on the PrivacyQA benchmark ~\cite{Ravichander2019Question}. We elevate the exiting baseline performance significantly (11\% F1) and achieve a new state-of-the-art result (50\% F1). Furthermore, our ablation studies provide an insightful understanding of our model.

% Second, privacy policies are ambiguous: even skilled annotators disagree on their interpretation for at least 26\% annotated policies in PrivacyQA benchmark. 
% % \notewa{Confusing, what point we are trying to prove here?}
%  We further propose a novel retriever ensemble technique to combat the second challenge
% %  \notewa{by the time readers reach here, they forget what was the second challenge}.
%  The idea is to use multiple pre-trained LMs with different pre-training objectives to build a set of retrievers and oracles to generate better augmented QA pairs. Our approach is motivated by the observation that retriever models built upon pre-trained models with different pre-training objectives learn different language representations and retrieve largely disjoint sets of candidates\notewa{this needs more clarification}. Figure~\ref{fig:model} illustrates our framework. Finally, we train a QA model on the augmented training set annotated by our framework. 
 
We evaluate our framework on the PrivacyQA benchmark. 
We elevate the state-of-the-art performance significantly (10\% F1) and achieve a new one (50\% F1). Furthermore, our ablation studies provide an insightful understanding of our model.
We will release all data and code upon acceptance.
% Code for reproducing experiments will be released upon acceptance.

\section{Methodology}
\label{sec:method}
% \notewa{Seems like our work is only focusing on a dataset, rather than a problem. Please, revise writing and explain the issue of imbalanced positive and negative answer sentences.}
% We first mitigate the data imbalance bottleneck in sentence selection/classification task by augmenting the positive examples in the training set. These positive examples, we retrieve from a large unlabeled corpus and filter out the noisy ones using an oracle model. We ensemble different retrievers and oracles built upon various pre-trained LMs. Next, we provide details of our modules and schema. 

% We need a bit more set up here. We should first discuss the original setup of the QA and then say our goals is to augment the training data. We also need to define the "positive" and "negative" data more directly somewhere.  

% Otherwise readers would confuse what are "quarries" and what are "policy sentenes" and what is the goal of the QA.

The privacy policy QA is a binary classification task that takes a user query $q$, a sentence $p$ from policy documents and outputs a binary label $z \in \{0,1\}$ that indicates if $q$ and $p$ are relevant or not. As most sentences $p$ are labeled as negative, our goal is to retrieve relevant sentences to augment the training data and mitigate the data imbalance issue. 
%Let us define a QA training dataset $D=\{(q_i, p_i, z_i)\}_{i=1}^m$, where $z_i \in \{0,1\}$ indicates whether $q_i$ and $p_i$ are relevant. 
Given a QA training dataset $D=\{(q_i, p_i, z_i)\}_{i=1}^m$, for each question in $D$, we (1) retrieve positive sentences from a large unlabeled corpus. (2) filter the noisy examples using filter models and aggregate final candidates. The final candidates are combined with the base data $D$ to train the end QA model. We use an ensemble of retrievers and filter models built upon various pre-trained LMs throughout the whole process. Details are discussed in the following.

% \smallskip\noindent
\paragraph{Retriever.}
% \paragraph{Retriever.}
\label{sec:method:ret}
Our retriever module is built upon the Dense Passage Retriever (DPR)  model~\cite{karpukhin2020dense}. 
It consists of two encoders $Q(\cdot)$ and $P(\cdot)$ that encode the queries and the policy sentences, respectively. 
% The similarity of a query $q$ and a policy sentence $p$ is defined by the dot product of their vector representations $\text{sim}(q, p)=Q(q)^T \cdot P(p)$.
The relevance of a query $q$ and a policy sentence $p$ is calculated by the dot product of $Q(q)$ and $P(p)$, i.e., $\text{sim}(q, p)=Q(q)^T \cdot P(p)$.
% Given a training set $D=\{(q_i, p_i, z_i)\}_{i=1}^m$, where $z_i \in \{0,1\}$ indicates whether $q_i$ and $p_i$ are relevant,
% For each positive pair in $D$, it optimizes only the cross-entropy loss with in-batch negatives~\cite{henderson2017efficient, parvez-etal-2021-retrieval-augmented}.
% <$q_i, p_i$> with $z_i=1$. 
% %%%%%%%%%%%%% Equation saving space %%%%%%%%%%%%%%%
% The training objective for <$q_i, p_i$> with a mini-batch of $n$ pairs is:\notewa{if space is an issue, omit this since everybody is familiar with DPR now}
% \begin{equation*}
% \setlength{\belowdisplayskip}{5pt}
% \setlength{\abovedisplayskip}{5pt}
% \label{eq:loss-ori-dpr}
%     \begin{split}
%     L_i = - \log \frac{e^{sim(q_i,p_i)}}{\sum_{j=1}^n e^{sim(q_i,p_j)}}.
%  \end{split}
% \end{equation*}
%%%%%%%%%%%%% Equation saving space %%%%%%%%%%%%%%%
% Denote the retriever with $\mathcal{R}_L$ that takes a pre-trained LM $L$ to initialize the encoders and an annotated training set, $D$. 
We train a retriever $\mathcal{R}_L$ on $D$, where the encoders in $\mathcal{R}_L$ are initialized with a pre-trained LM $L$.
% At inference, $\mathcal{R}_L$ takes an unlabeled corpus of policy sentences $\mathcal{P}=\{\mathcal{P}_1,\dots,\mathcal{P}_K\}$, and for each  query  $q \in D_q$ where $D_q = \{q_i: \forall (q_i, p_i, z_i) \in D\}$, it retrieves the top-$k$  most relevant policy sentences.
% Formally, the retriever output $\mathcal{R}_L(D,\mathcal{P}, k) =\{q,P_{j},1: j \leq k; \forall q \in D_q \}$  where each $P_{j} \in \mathcal{P}$ and $sim(q, P_{j}) \geq sim(q, P_{j+1}) \forall j$.
At inference, $\mathcal{R}_L$ retrieves the top-$k$  most relevant policy sentences from an unlabeled corpus of policy sentences $\mathcal{P}=\{p_1,\dots,p_M\}$ for each query $q_i$ in $D$, i.e., $\mathcal{R}_L(\{q_i\}_{i=1}^m,\mathcal{P}, k) =\{(q_i, p_{j}, 1): i\in[m], p_j \in \mathcal{P}_{\text{top}}(q_i, k)\}$, where $\mathcal{P}_{\text{top}}(q_i, k) \vcentcolon= \argmax_{\mathcal{P}' \subset \mathcal{P}, |\mathcal{P}'|=k} \sum_{p\in\mathcal{P}'} \text{sim}(q_i, p)$.

% \smallskip\noindent
\paragraph{Filtering Model.}
% \paragraph{Filtering Oracle.}
\label{sec:method:oracle} 
% To filter out the noisy retrievals from $\mathcal{R}_L(\{q_i\}_{i=1}^m,\mathcal{P}, k)$, we train a QA model, $\mathcal{Q}_L$. Essentially, it is a binary classification model that takes a query $q$ and a policy statement $p$ and predicts if they are relevant or not (i.e., $\mathcal{Q}_L(q,p) \in \{0,1\}$). 
To filter out the misclassified retrievals from $\mathcal{R}_L(\{q_i\}_{i=1}^m,\mathcal{P}, k)$, we train a QA (i.e., a  text-classification)  model ($\mathcal{Q}_L$) as a filter 
% . $\mathcal{Q}_L$ is a binary classification model
to predict whether a query $q$ and a (retrieved) policy sentence $p$ are relevant or not (i.e., $\mathcal{Q}_L(q,p) \in \{0,1\}$).
% using the training data $D$.
% Note that both $\mathcal{Q}_L$ and retriever $\mathcal{R}_L$ are built upon the same pre-trained LM $L$. 
% However, they only differ in training objectives (ranking problem vs binary classification; w/ and w/o in-batch negatives) and model architectures. 
Note that, the retriever model is a bi-encoder model that can pre-encode, index, and rank a large number of candidates. In contrast, our filtering model is a single cross-encoder model that can achieve comparatively higher performances~\cite{humeau2019poly} (i.e., hence better as a filter) but can not pre-encode and hence can not be used for large-scale retrieval (more differences are below and in Appendix  \ref{appendix:sec:diff-filter-dpr}).  Once the retrieval is done, $\mathcal{Q}_L$ serves as an inexpensive binary classifier which naturally suits as a filter and fit into the pipeline. 
% We treat one of them ($\mathcal{Q}_L$) as an oracle to filter the outputs of another one ($\mathcal{R}_L$) for label mismatches. 
We verify the effectiveness of our filtering in Section \ref{sec:ablation}.
We denote retrieval outputs after filtering as $\mathcal{D}_L = \{ (q,p,1) : \mathcal{Q}_L(q,p)=1, \forall (q,p,1) \in \mathcal{R}_L(\{q_i\}_{i=1}^m,\mathcal{P}, k) \}$. 

% \smallskip\noindent
\paragraph{Training of $\mathcal{Q}_L$ and $\mathcal{R}_L$.} Both the single encoder in $\mathcal{Q}_L$ and the two encoders in  $\mathcal{R}_L$ are initialized with the same pre-trained LM $L$ (e.g., BERT). Both are then fine-tuned as a binary classifier using the paired (query and policy is relevant or not) training data $D$. Additionally, to better-train the relatively weak bi-encoder $\mathcal{R}_L$, we consider the in-batch negative examples schema~\cite{henderson2017efficient, parvez-etal-2021-retrieval-augmented} and its hyper-parameters are tuned using mean ranking or mean reciprocal ranking (MRR) loss. At inference, for query $q$ and candidate $p$,  raw scores
from $\mathcal{R}_L$ is used to rank and prediction \{0,1\} from $\mathcal{Q}_L$ is used to filter $p$.

% \smallskip\noindent
\paragraph{Ensemble.}
% \paragraph{Ensemble.}
\label{sec:method:ensemble}  
% Unlike other domains, a privacy policy sentence can frequently have diverse interpretations (see Table \ref{table:qa-task}).
 % A single retrieved corpus $\mathcal{D}_L$ may not capture the necessary 
In order to enhance the diversity and the quality of the retrieved candidates, we use a set of pre-trained LMs $\mathcal{L} = \{L_1,\dots,L_l\}$ and aggregate all the corresponding retrieved corpora,
$\mathcal{D}_{\text{aug}} = \bigcup_{L\in \mathcal{L}} \mathcal{D}_{L}$.
In Section~\ref{sec:exp}, we show that retrieved corpora using multiple pre-trained LMs with different learning objectives can bring a different set of relevant candidates.
Lastly, we aggregate $\mathcal{D}_{\text{aug}}$ with $D$ (i.e., final train corpus $\mathcal{T} = \mathcal{D}_{\text{aug}} \cup D$) and train our final QA model with user specifications (e.g., architecture, pre-trained LM). 
% Therefore, we hypothesize that retrieved corpora using multiple different pre-trained LMs may bring different set of relevant candidates and in Section \ref{sec:exp}, we validate this. 

% \footnote{Note that, the union operation is used just to aggregate the train instances which does not hinder the final model training.} 
 
% \notewa{the motivation for ensemble looks weak}
% \notesc{$\mathcal{D}$ and $D$ are similar, How about adding a subscript, e.g., $\mathcal{D}_{\text{aug}}$}
\section{Experiments}
\label{sec:exp}

% 4) Which Pretrained model to use in augmentation 
% 8) Page-1 example 
% 13) Retrieval example

% In this section, we evaluate our approach and present the findings from our analysis.

% \input{table/data_stat}
% \input{table/main_result.tex}

\subsection{Setup} 

% \paragraph{Implementation.}
% \smallskip\noindent
\paragraph{Evaluation Metrics.} 
% \paragraph{Settings} 
% First we describe the settings.  Our final goal is to perform QA task for privacy domain. As mentioned before this is in fact a text classification task and we evaluate on PrivacyQA benchmark. 
We evaluate our approach on PrivacyQA that is framed as a text classification task \cite{Ravichander2019Question}.
We use \emph{precision}, \emph{recall}, and \emph{F1 score} 
as the evaluation metrics.

\paragraph{Implementations.} 
% to evaluate the model performance. 
As for the retrieval database $\mathcal{P}$, we crawl privacy policies from the most popular mobile apps spanning different app categories in the Google Play Store and end up with 6.5k documents (0.6M statements). By default, all retrievals use top-$10$ candidates w/o filtering.
% \notemrp{Jianfeng, cd u please give a short description ana a data stat of our crawled dataset. }. 
All data/models/codes are implemented using (i) Huggingface Transformers~\cite{Wolf2019HuggingFacesTS}, (ii) DPR~\cite{karpukhin2020dense} libraries.
% and will be released. 

\begin{table}[!t]
\centering
% \scriptsize
% \hspace{-1.7mm}
\resizebox{\linewidth}{!}{
\begin{tabular}{@{}l@{\hskip 0.02in}|c@{\hskip 0.05in}|c@{\hskip 0.05in}|c@{\hskip 0.05in}|c@{}} 
\hline
Method & F & Precision & Recall & F1 \\ 
\hline\hline
Human & - & $68.8$ & $69.0$ & $68.9$  \\ \hline
\multicolumn{5}{@{}l}{W/o data augmentation} \\
\hline
% \multirowcell{2}{Fine- \\ tuning} & \multirowcell{2}{SQuAD \\ Pre-training}  & \multicolumn{2}{c|}{Valid} & \multicolumn{2}{c}{Test}  \\
% \cline{3-6}
BERT+Unans. & \multirowcell{4}{-} & $44.3$ & $36.9$ & $39.8$ \\
BERT (reprod)  &  & $48.0_{\pm 2.0}$ & $37.7_{\pm 1.2}$ &	$42.2_{\pm 1.5}$  \\
% \emph{PBERT}  &  & 50.8	& 43.1 &	46.7 \\
PBERT  &  & $51.2_{\pm 0.4}$ & $42.7_{\pm 0.6}$ &	$46.6_{\pm 0.4}$ \\
% \emph{SimCSE}  &  & 48.8 &	42.2 &	45.3\\
SimCSE  &  & $48.4_{\pm 0.8}$ &	$41.4_{\pm 0.7}$ &	$44.7_{\pm 0.7}$ \\

% \emph{SimCSE} (subsample 7137 pos 7137 neg) & 22.2  & 64.7 & 33.0 \\

\hline
\multicolumn{5}{@{}l}{Retriever augmented} \\
\hline
% \multirow{2}{*}{\emph{BERT-R}} & \xmark & 39.9 & 50.8 & 44.7 \\
\multirow{2}{*}{\emph{BERT-R}} & \xmark & $39.0_{\pm 0.8}$ & $52.4_{\pm 1.7}$ & $44.7_{\pm 0.4}$ \\
% & \cmark & 46.5 & 45.5 & 46.0 \\
& \cmark & $48.1_{\pm 1.4}$ & $44.7_{\pm 0.9}$ & $46.3_{\pm 0.5}$ \\
\hdashline
% \multirow{2}{*}{\emph{PBERT-R}} & \xmark & 48.4 & 45.6 & 46.9\\ 48.7 46.1
\multirow{2}{*}{\emph{PBERT-R}} & \xmark & $48.7_{\pm 1.9}$ & $44.1_{\pm 1.8}$ & $46.3_{\pm 1.6}$\\
% & \cmark & 49.5	& 46.3 &	47.8 \\
% & \cmark & $47.8_{\pm 1.48}$	& $44.9_{\pm 2.54}$ &	$46.3_{\pm 1.87}$ \\ 
& \cmark & $49.2_{\pm 1.6}$	& $44.9_{\pm 2.0}$ &	$47.0_{\pm 1.2}$ \\ 

% 46.85, 46.18, 46.12, 48.72 # F1 pbert filter
% 50.9, 49.8, 49.3,47.2 # prec  pbert filter
% 46.5, 43.4, 43, 46.7 # rec  pbert filter
\hdashline
% \multirow{2}{*}{\emph{SimCSE-R}} & \xmark & 48.4 & 47.2 & 47.8\\
\multirow{2}{*}{\emph{SimCSE-R}} & \xmark & $47.0_{\pm 2.1}$ & $44.5_{\pm 2.4}$ & $45.7_{\pm 1.9}$ \\
% & \cmark &  51.0 & 45.2 & 47.9 \\
& \cmark &  $48.6_{\pm 2.2}$ & $43.9_{\pm 1.2}$ & $46.1_{\pm 1.6}$ \\
\hline
\multicolumn{5}{@{}l}{Ensemble retriever augmented} \\
\hline
Baseline-E & \xmark & $22.2_{\pm 0.8}$ & $54.4_{\pm 0.8}$  & $31.4_{\pm 0.8}$ \\
% 32.3, 31.2, 30.8 \\
% \tool & \cmark & 47.1 & 52.9 & {\bf 49.8} \\
\tool & \cmark & $47.4_{\pm 0.6}$ & $50.5_{\pm 2.2}$ & ${\bf 48.9}_{\pm 0.8}$ \\
% Ours-PBERT & 51 & 45.9 & 48.3 \\
%  \toolextnospace & \cmark & 51.3 & 50.0 & {\bf 50.6} \\
\toolextnospace & \cmark & $51.0_{\pm 0.4}$ & $48.7_{\pm 0.9}$ & ${\bf 49.8}_{\pm 0.7}$ \\
\hline
\end{tabular}
}
% \vspace{-1mm}
\caption{
% \small
Test performances on PrivacyQA ($\text{mean}_{\pm \text{std}}$). F indicates filtering and BERT+Unans. refers to the previous SOTA performance~\cite{Ravichander2019Question}. Retrieved candidates improve all the baseline QA models, especially when being filtered. Our ensemble retriever approach combines them and achieves the highest gains. 
% \notewa{updated the table, please check}
}
\label{table:main-result}
% \vspace{-4mm}
\end{table}

% \smallskip\noindent
\paragraph{Baselines.}
% \paragraph{Baselines} 
We fine-tune three pre-trained LMs on PrivacyQA as baselines: (i) \emph{BERT}: Our first baseline is BERT-base-uncased ~\cite{devlin2019bert} which is pre-trained on generic NLP textual data. A previous implementation achieves the existing state-of-the-art performance (BERT+Unams. in \citet{Ravichander2019Question}). (ii) \emph{PBERT}: We adapt \emph{BERT} to the privacy domain by fine-tuning it using masked language modeling on a corpus of 130k privacy policies (137M words) collected from apps in the Google Play Store~\citep{harkous2018polisis}. Note that the retrieval database $\mathcal{P}$ is a subset of this data that is less noisy and crawled as a recent snapshot (more in Appendix \ref{appendix:diff-corpora})
(iii) \emph{SimCSE}: We take the \emph{PBERT} model and apply the unsupervised contrasting learning SimCSE~\cite{gao-etal-2021-simcse} model on the same 130k privacy policy corpus. 
% Note that we train (fine-tune) all three baselines as final QA models on PrivacyQA.
We also consider three other retrieval augmented QA models based on individual pre-trained LM without ensemble: (iv) \emph{BERT-R}: $\mathcal{L}=\{\emph{BERT}\}$, (v) \emph{PBERT-R}: $\mathcal{L}=\{\emph{PBERT}\}$, (vi) \emph{SimCSE-R}: $\mathcal{L}=\{ \emph{SimCSE}\}$. 
We first construct $\mathcal{T}$ (both settings: w/ and w/o filter model) and  fine-tune on it the corresponding pre-trained LM as the final QA model. Finally, we consider one more ensemble retrieval augmented baseline (vii) \emph{Baseline-E}, which is precisely the same as ours (settings below), except there are no intermediate filtering models.

% \smallskip\noindent
\paragraph{Ours.}
We construct the augmented corpus $\mathcal{T}$
% , discussed in Section \ref{sec:method:ensemble}, 
using (i) all 3 aforementioned pre-trained LMs: $\mathcal{L} = \{$\emph{BERT}, \emph{PBERT},  \emph{SimCSE}$\}$ (ii) domain adapted models only: $\mathcal{L} = \{$\emph{PBERT},  \emph{SimCSE}$\}$. For brevity, we call them: \emph{Ensemble Retriever Augmentation} (\toolnospace) and \emph{Ensemble Retriever Augmentation--Domain Adapted} (\toolextnospace).
By default, we fine-tune \emph{SimCSE} as the final QA model.

\subsection{Main Results}
% \paragraph{Results and Analysis} 
The results are reported in Table \ref{table:main-result}. Overall, domain adapted models \emph{PBERT} and \emph{SimCSE} excel better than the generic \emph{BERT} model.
The retrieval-augmented models enhance the performances more, especially the recall score, as they are added as additional positive examples. 
% However, they might contain several noisy examples (see Appendix \ref{appendix:sec:qual-example}), and filtering those out improves the precision scores for all three retrievers.
However, these models may contain noisy examples (see Table \ref{table:ret-example-details-main}), which lowers precision. Filtering these examples leads to improved precision for all retrievers. 
Finally, \tool and \toolext aggregate these high-quality filtered policies--leading toward the highest gain (10\% F1 from the previous baseline) and a new state-of-the-art result with an F1 score of $\sim$50. Note that \emph{Baseline-E} unifies all the candidates w/o any filtering performs considerably worse than all other models, including each individual retrieval model: \emph{Baseline-E} augments more candidates as positives, which explains the highest recall score; in the meantime, as it does not filter any, the corresponding precision score is oppositely the lowest.

\begin{table}[t]
\centering
% \scriptsize
% \hspace{-2.2mm}
% \small
% \resizebox{1.0\linewidth}{!}{
\begin{tabular}{@{\hskip 0.05in}l@{\hskip 0.05in}|c@{\hskip 0.05in}|c@{\hskip 0.05in}|c@{\hskip 0.05in}|c@{\hskip 0.05in}|c@{\hskip 0.05in}} 
\hline
Query Type & \% & B & PB & S & \toolnospace \\
\hline\hline
Data Collection & 42 & 45 & {\bf 46} & {\bf 46} & {\bf {\color{blue} 48}} \\
% Data Collection & 42 & 45 & {\bf 46} & {\bf 45} &{\color{blue} \bf 47.2} \\
Data Sharing & 25 &  {\bf 43}  & {37} & {41} & {\color{blue} \bf 43} \\ 
% Data Sharing & 25 &  {\bf 43}  & {37} & {36.1} & {\color{blue} \bf 43.2} \\ 
Data Security & 11 & {\bf 65} & {61} & 60 & 60 \\ 
% Data Security & 11 & {\bf 65} & {61} & 60 & 60.6 \\ 
% Data Retention  & 4 & {\bf 52/102}  & {35/95}  & 35/97 & {\color{blue} \bf 56/173} \\ 
Data Retention & 4 & {\bf 52}  & {35}  & 35 & {\color{blue} \bf 56} \\  
User Access  & 2 &  {\bf 72}  & 48  & 31 & 61 \\
% User Access & 2 &  {\bf 72}  & 48  & 31 & 57.8 \\
User Choice & 7 & 41 & {\bf 60} & 42 & {\color{red} 31} \\
% User Choice & 7 & 41 & {\bf 60} & 42 & {\color{red} 29.9}\\
Others & 9 &  {36} & {{45}} & {\bf 52} & {\color{blue} \bf 55} \\ 
% Others & 9 &  {36} & {{45}} & {\bf 52} & {\color{blue} \bf 52.3} \\ 
\hline
Overall & 100 & 45 & 47 & 48 &  {\color{blue} \bf  49} \\
\hline 
% \# correct (total) & - & 2265 &  2096 & 1988 & 2935 \\
% Overall & 100 & 45 & 47 & 48 &  {\color{blue} \bf  48.9} \\
\hline
\end{tabular}
% }
% \vspace{-1mm}
\caption{
% \small
F1-score breakdown (values are in Appendix \ref{appendix:sec:privacyqa}). B, PB, S refers to retrievers \emph{BERT-R}, \emph{PBERT-R}, and \emph{SimCSE-R}. 
%T, N refers to temporal and numeric questions. 
Different models performs better for different types (black-bold). \tool combines them and enhances performances for all categories (except: red).} 
\label{table:breakdown_result}
% \vspace{-6mm}
\end{table}

%\textbf{Analysis.} 
\subsection{Analysis}
Table~\ref{table:breakdown_result} shows the  performance breakdown for different query types (more in Appendix~\ref{appendix:sec:privacyqa}).
For questions related to data collection, data sharing, and data security, the performance difference among the models is relatively small ($\leq$ 5\% F1); for data retention and user access, \emph{BERT-R}, that is pre-trained on generic NLP texts, performs significantly well ($>$ 15\% F1), possibly because the answers to these query types focus on providing numerical evidence for the questions (e.g., How many days the data are retained?) that is less relevant to the domain of privacy policies; and for other types of questions the domain adapted models performs better  ($>$ 15\% F1). Overall, the individual retrieval augmented models based on LM pre-trained w/ different corpora and objectives perform at different scales for each type, and combining their expertise, \tool enhances the performances for all types. 

\begin{figure}[t]
\captionsetup[subfigure]{labelformat=empty}
% \vspace{-1mm}
\centering
% \hspace{-8mm}
\subfloat[(a) \label{fig:qtype_anal}]
{
\includegraphics[width=0.22\textwidth]{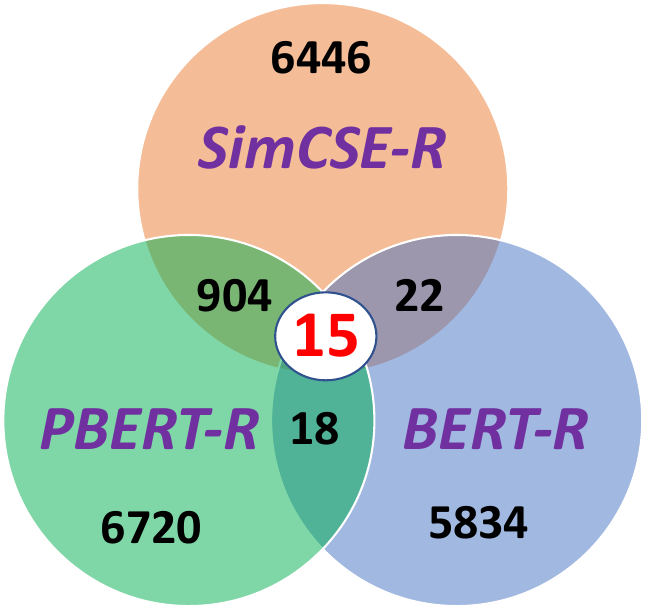}
%\vspace{-3mm}
}
% &
\hfil
% \hspace{-3mm}
\subfloat[(b) \label{fig:atype_anal}]
{
\includegraphics[width=0.22\textwidth]{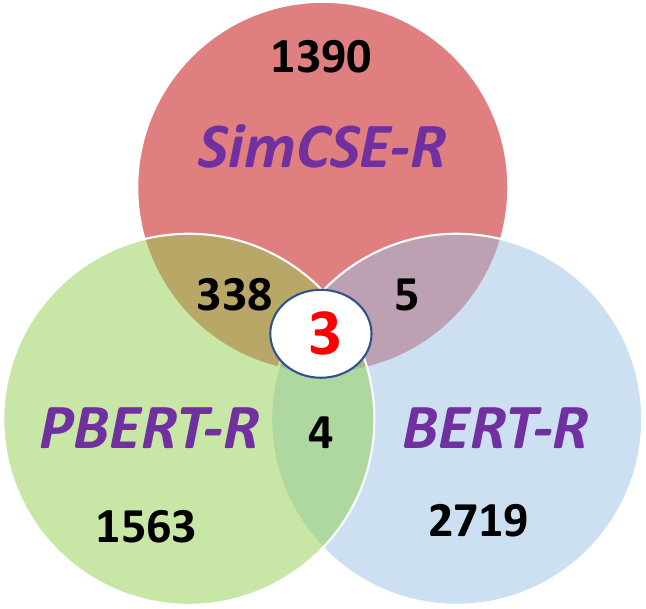}
% \vspace{-3mm}
}
% \vspace{-1mm}
\caption{ Venn diagram of low mutual agreement (<1\%) among retrievers (a); even amplified after filtering (b). 
}
% \vspace{-2mm}
\label{figure:venn}
\end{figure}

Next, we show the Venn diagram of overlapping retrievals in Figure~\ref{figure:venn}. Although policy statements describe similar information (i.e., have common phrases), they are often verbose and equivocal (i.e., multiple-different interpretations). Consequently, retrievers w/ different objectives and training corpora rank them differently. Therefore, although being retrieved from the same corpus, candidates retrieved by different models rarely match fully but may have notable overlapping information (words/phrases) and improve their  performances equitably. For example, while the performances of  \emph{BERT-R}, \emph{PBERT-R} and \emph{SimCSE-R} w/ filtering are similar ($\sim$46) in Table \ref{table:main-result}, from Figure~\ref{figure:venn} their overlapping (exact match) is $<$ 1\% (qualitative examples in Appendix \ref{appendix:sec:qual-example}). At the same time, their raw retrieval corpora have a high BLEU score of ($\geq 0.78$). This validates our hypothesis that retrievers built upon different pre-trained LMs learn distinct representations and hence retrieve diverse candidates.

% \begin{figure}[t]
% % \centering  
% \hspace{-5pt}
% \includegraphics[width=0.2\textwidth]{images/Vend-filtered.pdf}
% \caption{\footnotesize { }
% \vspace{-0.2cm}
% }
%   \label{fig:model}
% \end{figure}

% \begin{figure}[t]
% \centering
% % \hspace{-5pt}
% \includegraphics[width=0.2\textwidth]{images/Vend-unfiltered.pdf}
% \caption{\footnotesize {
% %  \vspace{1cm}
% Venn diagram of low mutual agreement (<1\%) among retrievers  that  gets even amplified after filtering. }
% % \vspace{-0.5cm}
% }
%   \label{fig:model}
% \end{figure}

\subsection{Ablation Study}
\label{sec:ablation}

\paragraph{Sampling to tackle data imbalance.} 
The preliminary experiments studied rebalancing techniques like equal sampling, but oversampling does not add new information and undersampling limits data, leading to poor generalization on unseen test data.  Using equal positive and negative sub-sampled training instances resulted in a 9\% drop in F1 score. Even augmented with a higher number of filtered positive examples retrieved by a single retriever model does not perform as well as when a lower number of ensemble-based positive instances are augmented. From Table \ref{table:filter-impact-appendix} in Appendix, augmenting with a high number of filtered positive examples from a single retriever model performed worse than using a lower number of ensemble-based positive examples--suggesting the need for diverse and high-quality knowledge not present in training data. 
% (More in Appendix \ref{appendix:more-ablation}).

% \smallskip\noindent

\paragraph{A common filter.} Performances of \tool (last row in Table~\ref{table:breakdown_result}) with a common filter model based on \emph{SimCSE} for all the retrievers regardless of their corresponding pre-trained models are 49.2, 45.2, and 47.1, respectively--validating the requirement of filtering using the corresponding pre-trained LM.  
% \smallskip\noindent

\paragraph{Other pre-trained LM as the final QA model.} Fine-tuning \emph{PBERT} instead of \emph{SimCSE} on  $\mathcal{T}$ (last two rows in Table~\ref{table:main-result}) becomes:
47.0, 47.1, 47.0 and 51.0, 45.9, 48.3, respectively--showing that \tool is generic to end model choices.

% \smallskip\noindent
\paragraph{Which pre-trained LMs to use?} 
Table~\ref{table:breakdown_result} shows
\toolext that combines fewer pre-trained LMs can outperform the one with more models, \toolnospace. Though here we consider a simple approach (in-domain) for selecting the potential subset of models, this paves a new direction for future research (e.g., \citet{parvez2021evaluating}).
% More ablation studies are in the Appendix \ref{appendix:more-ablation}. 

\begin{table}[!ht]
\centering
% \vspace{4mm}
\resizebox{\linewidth}{!}{%
% \small
\def\arraystretch{1.5}%
\begin{tabular}{p{0.99\linewidth}}
\hline
Q: do you sell my photos to anyone? \\
\hline
% \medskip
\vspace{-6mm}
{\bf Gold:} i) We use third-parties to serve ads on our behalf across the Internet. 
% (ii)
% These companies may use your personal information to enhance and personalize your shopping experience with us, to communicate with you about products and events that may be of interest to you and for other promotional purposes. 
%  iii) Your use of our Application with that healthcare institution may be subject to that healthcare institution's policies and terms. 
 (ii) We may share personal information within our family of brands. 
(iii) From time to time we share the personal information we collect with trusted companies who work with or on our behalf.
(iv) No personally identifiable information is collected in this process.
\\
\vspace{-6mm}
% \noskip
{\color{blue}\bf Correct Retrievals:} 
(i) \emph{SimCSE-R}: The Application does not collect or transmit personally identifiable information such as your name, address, phone number or email address.
% (ii) Some of this information is automatically gathered, and could be considered personally identifiable in certain circumstances, however it will generally always be anonymised prior to being viewed by Not Doppler, and never sold or shared.
% -(BERT-R)
(ii) \emph{PBERT-R}: We also use the Google AdWords to serve ads on our behalf across the Internet and sometimes on this Website.
(iii) \emph{BERT-R}: To organ and tissue donation requests: By law, we can disclose your health information  to organ procurement organizations.
\\
\vspace{-6mm}
{\color{red}\bf Incorrect Retrievals:} 
(i) \emph{BERT-R}: These are not linked to any information that is personally identifiable. (ii) \emph{SimCSE-R}: When you upload photos to our platform or give us permission to access the photos on your device, your photo content may also include related information such as the time and place your photo was taken and similar “metadata” captured by your image capture device. 
\\
\hline
\end{tabular}
}
% \vspace{-2mm}
\caption{Example retrieved policies. Retrieved candidates are distinct from expert annotated ones and can bring auxiliary knowledge to the model. Filtering is needed as inappropriate candidates can also be retrieved. }
\label{table:ret-example-details-main}
% \vspace{-6mm}
\end{table}

% \smallskip\noindent
% \textbf{Qualitative examples.} Table \ref{table:ret-example-} (more in Appendix) shows some example retrievals of different models. They are distinct from expert annotated ones and can bring auxiliary knowledge.  

\paragraph{Recall performances on PrivacyQA dataset.} Example retriever \emph{BERT-R} scores the recall@k (k up to 10) values as 17, 28, 36, 42, 48, 53, 58, 59, 63, 67 respectively while the (cross-encoder) BERT QA model (i.e., filter model) achieves a recall (same as recall@1) of $\sim37$. This shows the effectiveness of our designed filter model.

\paragraph{Can the final QA model be used as a filter and impact of filter models on the end performance?} 
The final QA model can be used as a filter model. As for the single retriever model-based augmentation, the performance of the downstream end task depends on the performance of the retriever model, the filtering model, and the end QA model. 
A stronger filter model leads to better end QA performance in general.  With a \emph{BERT-R} retriever and \emph{BERT} end QA model, the use of a \emph{PrivacyBERT} filter model improves that of a \emph{BERT} filter model from 46.3 to 46.8. However, when doing the ensembling using a combination of retriever and filter models from the same pre-trained language model (PLM) results in even better performance than using a stronger filter model from different PLMs.
% (more in Appendix  \ref{appendix:more-ablation})

\paragraph{Qualitative examples.}
Table \ref{table:ret-example-details-main} shows some example retrievals of different models. Retrieved candidates are distinct from expert annotated ones and can bring auxiliary knowledge to the model. 

\section{Conclusion}
\label{sec:conclusion}
% We develop a data augmentation framework for QA  (sentence selection) task on privacy policies.
% \smallskip\noindent
% \textbf{Conclusion}
We develop a noise-reduced retrieval-based data augmentation method that combines different pre-trained language models to address the data imbalance issue in privacy policy QA.
% Although we focus on a specific domain,
However,
our approach can possibly be adapted to other domains and we leave the exploration as the future work.
% Our framework first builds a set of retrievers using different pre-trained LMs to retrieve positives policies from an unlabelled corpus. We then filter the retrievals using an oracle model that is also learned using the corresponding pre-trained LMs. Finally we aggregate them and train our end QA model. 

% , such as legal NLP~\cite{zhong-etal-2020-nlp}.

\section*{Limitations}

% { \color{red} We should add limitations here. Otherwise, the paper will be desk-rejected.}
In this work, we develop a retrieval-augmented QA framework specifically for Privacy Policies. Its effectiveness rooted on the characteristics of disjoint retrievals from different pre-trained language models (PLMs). Although this focused work completely aligns, addresses and adheres to the guidelines for a short-paper in this venue,  we have not performed any experiments on data outside this privacy policy domain. Hence, the applicability of our method for any generic data domain is unknown. While the time/latency and resource utilization remains unchanged at inference-time, using multiple retriever modules our method introduce an additional overhead in model training. 
% The mean and standard deviations of our results are computed for no more than 5 runs with random seeds (1, 2, 3, 4, 5, 10 and the default seed of Huggingface Transformers Library\footnote{\url{https://github.com/huggingface/transformers}}). 
Using different implementation, PLMs, and random seeds may also lead to results that could be different from ours.

\section*{Ethics Statement}
In this work, our approach crawls an unlabeled privacy policy corpus from the web policy documents specifically from the Google play store which we use as a retrieval database. Although these documents are completely publicly available and was used only for research purpose, they may contain some nomenclature of certain persons, objects, products, users, developers, or production houses (i.e., industries). We neither obfuscate nor make any altercation/modification to them. 

\section*{Acknowledgments}
We thank the anonymous reviewers for their insightful comments. This work was supported in part by National Science Foundation Grant OAC 2002985, OAC 1920462 and CNS 1943100, Google Research Award, and Meta Research Award. Any opinions, findings, conclusions, or recommendations expressed herein are those of the authors and do not necessarily reflect those of the US Government or NSF.
% Entries for the entire Anthology, followed by custom entries
\bibliography{references/policyIE, references/policyqa, references/berteer}
\bibliographystyle{acl_natbib}

\clearpage
\appendix

\twocolumn[{%
 \centering
 \Large\bf Supplementary Material: Appendices \\ [15pt]
}]

% \section{Discussion and Conclusion}
\section{Related  Works}
%\label{sec:related-work}

% Briefly mention some work in privacy policies
% Talk about the recent progress on QA for privacy policies (what types of datasets are available)
% implication and connection to QA tasks in other domain
% \smallskip\noindent
% \textbf{Related Works}
A line of works focuses on using NLP techniques for privacy policies~\cite{wilson2016creation, harkous2018polisis, zimmeck2019maps, bui2021automated, ahmad2021intent}.
Besides the QA tasks as sentence selection, \citet{ahmad2020policyqa} propose another SQuAD-like~\cite{rajpurkar2016squad} privacy policy reading comprehension dataset  for a limited number of queries. 
Oppositely, we focus on the more challenging one, which allows unanswerable questions and ``non-contiguous'' answer~\cite{ravichander-etal-2021-breaking}. 
In relevant literature works, retrieval augmented methods are applied in various contexts including privacy policies (e.g., \citet{van2021cheap,keymanesh2021privacy,yang2020neural}). 
Non-retrieval data aggregation has also been studied under different
NLP contexts (e.g., bagging \cite{breiman1996bagging}, meta learning \cite{parvez2019robust}).
However, we uniquely aggregate the retriever outputs using different pre-trained language models.
% sentence outputs from retrievers 

% \notesc{Talk about the implication and connection of our works to QA tasks in other domains, e.g., whether our approach could be extended to other domains, etc.}

\section{Limitations/Reproduction}
% We develop a retrieval based data augmentation framework on  that uses the combination of different pre-trained LMs as a backbone.
In this paper, we show that leveraging multiple different pre-trained LMs can augment high-quality training examples
and enhance the QA (sentence selection) task on privacy policies. Our approach is generic and such unification of different kinds of pre-trained language models for text data augmentation can improve many other low-resourced tasks or domains. However, it is possible that our approach:

\begin{itemize}
    \item may not work well on other scenarios (e.g., domains/language or tasks etc.,).
    \item subject to the choice of a particular set of models. For example, as mentioned in Section, \ref{sec:ablation}, fine-tuning  pre-trained models other than \emph{SimCSE}~\cite{gao-etal-2021-simcse} as the final QA model  achieve lower gain.
    \item may not work for certain top-$k$ retrievals. For example, from Table \ref{table:filter-impact-appendix}, we get different results with different scales for variable top-$k$ values (e.g., top-10, top-100).  
        \item uses the same set of hyperparameters for all:
        \begin{itemize}
            \item QA model: 
            \begin{itemize}
                \item learning rate: 2$e^{-5}$,
                \item train epoch: 4,
                \item per gpu train batch size: 31,
                \item num gpus: 4
                \item fp16 enabled
                \item others: mostly default as in Huggingface 
                \item train time: around 2 hours
                \item Higgingface transformer version 0.3.2. (it has Apache License 2.0)
            \end{itemize}
        \end{itemize}
        \begin{itemize}
            \item Retriever model: 
            \begin{itemize}
                \item learning rate: 2$e^{-5}$,
                \item train batch size: 16,
                \item train epoch: 100,
                \item global\_loss\_buf\_sz 600000,
                \item others: mostly default as in DPR  (It has  Attribution-NonCommercial 4.0 International license)
                \item num gpus: 3
                \item Higgingface transformer version 0.3.2 (it has Apache License 2.0)
                \item train time: around 12-18 hours
            \end{itemize}
        \end{itemize}
\end{itemize}

As our primary goal is on the retrieval-based data augmentation
technique, we expect further optimization of task-specific model hyperparameters to improve performance. Note that our results are based on upto 4 runs using different random seeds.
% (3 for all , 4 for PBERT-r filter)
%and running it multiple times with different random seeds may incur slight variation from the results we report. 

% \input{latex/ablation}

\section{Privacy Policy Data Crawling \& Retrieval Statistics}
% Did you discuss the steps taken to check whether the data that was collected/used contains
% any information that names or uniquely identifies individual people or offensive content,
% and the steps taken to protect / anonymize it?
We crawl our English retrieval corpus from Google App Store using the Play Store Scraper\footnote{https://github.com/danieliu/play-scraper}.
% The data we crawled is about privacy policies. In general, a privacy policy does not contain any personally identifiable information. However, there could be some mention of specific nomenclatures. There is no easy way to remove them, so we did not filter them manually. Note that we do not intend to use any commercial usage. 

% \section{Retrieval Statistics per OPP-115~\cite{wilson2016creation} Query Type:}

\begin{table}[h]
\centering
% \scriptsize
% \hspace{-1.7mm}
% \small
% \resizebox{\linewidth}{!}{
\begin{tabular}{c|c} 
\hline
Query Type & No. of Retrieval \\
\hline\hline
Data Collection & 2893 \\
Data Sharing & 1848 \\
Data Security & 891 \\
Data Retention & 542 \\
User Access & 145 \\
User Choice & 335 \\
Others & 14 \\
\hline
\end{tabular}
% }
% \vspace{-1mm}
\caption{Retrieval statistics per query type.}
\label{table:query_stats}
% \vspace{-2mm}
\end{table}

However, below is the statistics of our (\toolnospace) augmented corpus per each question category in the PrivacyQA training set.

% \begin{itemize}
%     \item Q\_type:  Data Collection ;
%             Retrievals:  2893
%     \item Q\_type:  Data Sharing ;
%     Retrievals:  1848
%     \item Q\_type:  Data Security ;
%     Retrievals:  891
%     \item Q\_type:  Data Retention ;
%     Retrievals:  542
%     \item Q\_type:  User Access ;
%     Retrievals:  145
%     \item Q\_type:  User Choice ;
%     Retrievals:  335
%     \item Q\_type:  Others ;
%     Retrievals:  14
%     \item Q\_type: Specific Audiences ;
%     Retrievals:  14

% \end{itemize}

\section{PrivacyQA Dataset and Breakdown of Performance in Absolute Numbers}
\label{appendix:sec:privacyqa}

\begin{table*}[th]
\centering
% \scriptsize
% \hspace{-2.2mm}
% \small
% \resizebox{1.0\linewidth}{!}{
\begin{tabular}{@{\hskip 0.05in}l@{\hskip 0.05in}|c@{\hskip 0.05in}|c@{\hskip 0.05in}|c@{\hskip 0.05in}|c@{\hskip 0.05in}|c@{\hskip 0.05in}} 
\hline
Query Type & total & B & PB & S & \toolnospace \\
\hline\hline
Data Collection & 6280 & 1901 & {1157} & {1186} & {1806} \\
% Data Collection & 42 & 45 & {\bf 46} & {\bf 45} &{\color{blue} \bf 47.2} \\
Data Sharing & 4734 &  1332  & {777} & 1092  & {1268 } \\ 
% Data Sharing & 25 &  {\bf 43}  & {37} & {36.1} & {\color{blue} \bf 43.2} \\ 
Data Security & 994 & {416} & {399} & 423 & 393 \\ 
% Data Security & 11 & {\bf 65} & {61} & 60 & 60.6 \\ 
Data Retention  & 453 & {150}  & {98}  & 110  & 173 \\ 
% Data Retention & 4 & {\bf 52}  & {35}  & 35 & {\color{blue} \bf 54.2} \\  
User Access  & 221 &  89  &  47  & 43 &  87\\
% User Access & 2 &  {\bf 72}  & 48  & 31 & 57.8 \\
User Choice & 493 & 91 &  49 & 24 & 55 \\
% User Choice & 7 & 41 & {\bf 60} & 42 & {\color{red} 29.9}\\
Others & 28 &  2 & {{1}} & 2 & 4 \\ 
% Others & 9 &  {36} & {{45}} & {\bf 52} & {\color{blue} \bf 52.3} \\ 
\hline
Overall & 10332 & 3135 & 2084 & 2334 &  2935  \\
\hline 
% \# correct (total) & - & 2265 &  2096 & 1988 & 2935 \\
% Overall & 100 & 45 & 47 & 48 &  {\color{blue} \bf  48.9} \\
\hline
\end{tabular}
% }
% \vspace{-1mm}
\caption{
Number of correct predictions. Note that F1-score is not proportional to the accuracy.  B, PB, S refers to retrievers \emph{BERT-R}, \emph{PBERT-R}, and \emph{SimCSE-R}. 
%T, N refers to temporal and numeric questions. 
% Different models performs better for different types (black-bold). \tool combines them and enhances performances for all categories, in general (except: red).
}
\label{table:breakdown_result-ii}
% \vspace{-2mm}
\end{table*}
\begin{table}[!ht]
\centering
% \resizebox{\linewidth}{!}{
% \small
\begin{tabular}{l|l}
\hline
& PrivacyQA \\
\hline
\hline
Source & Mobile application   \\
Question annotator & Mechanical Turkers \\
Form of QA & Sentence selection \\
Answer type  & A list of sentences \\
% \# Annotations  & 3,500 \\
\# Unique policy docs &  train: 27, test:  8 \\
\# Unique questions & train: 1350, test: 400 \\
% \# Sentences & train: 3704, test: 1243  \\
% \# QA instances &  train: 185201, test: 62151 \\
\# QA instances &  train: 185k, test: 10k \\
Avg Q. Length &  train: 8.42 test: 8.56 \\
% Avg Doc. Length & train:  3121.3, test: 3629.13 \\
Avg Doc. Length & train:  3.1k, test: 3.6k \\
% Avg Ans. Length & train: 123.73, test: 153.44 \\
Avg Ans. Length & train: 124, test: 153 \\
\hline
\end{tabular}
% }
% \vspace{-2mm}
\caption{Brief summary of PrivacyQA.}
\label{table:privacyqa_vs_policyqa}
% \vspace{-2mm}
\end{table}

% \begin{tabular}[c]{@{}l@{}}Question Source \\ from Domain Experts \end{tabular}

% \section{Breakdown of Performance}
% \label{appendix:sec:breadwon-f1-score}

Table \ref{table:breakdown_result-ii} shows the accuracy breakdowns in absolute numbers. In addition, 
A brief summary of the PrivacyQA benchmark is in Table~\ref{table:privacyqa_vs_policyqa}. Note that the questions in Privacy can be categorized into a few OPP-115 classes. These categories  are enlisted in Table \ref{table:breakdown_result} in the main paper and the details of each category can be found in \citet{wilson2016creation}.

% \newpage

\section{Difference Between Pre-training and Retrieval Corpus}
\label{appendix:diff-corpora}

130k documents were collected before 2018 and by that time, the GDPR\footnote{https://gdpr-info.eu/} and CCPA\footnote{https://oag.ca.gov/privacy/ccpa} were not enforced by then. Thus, the 130k documents are out-of-date and some content might not be comprehensive as the retrieval corpus. Besides, the 130k documents provided by (Harkous et al., 2018) contains some noises since we observe that the documents are not all written in English. However, as the data size is larger, we still use it for pre-training. In contrast, our corpus was collected after 2020 and we filtered out some possible noises (e.g., filtering out non-English document) while crawling.

\section{Difference Between the Filtering Model and the Retriever}
\label{appendix:sec:diff-filter-dpr}
The retriever model is a bi-encoder model whose model parameters are fine-tuned with in-batch negative loss (discussed in Section \ref{sec:method:ret} in the main paper), hyper-parameters are tuned based on average rankings (\url{https://github.com/facebookresearch/DPR/blob/a31212dc0a54dfa85d8bfa01e1669f149ac832b7/train_dense_encoder.py#L294}) and that can pre-encode, index and rank a large number of candidates while our filtering model is a cross-encoder text-classifier (e.g., single encoder fine-tuned BERT) that is fine-tuned w/o any additional in-batch negatives and in-general achieves comparatively higher performance~\cite{humeau2019poly} (i.e, better as a filter) but can not pre-encode and hence can not be used for large scale retrieval.

% \newpage 
% \section{Effectiveness of Oracle Filtering and Different Top-$k$ Selection}

\begin{table}[h]
\centering
% \scriptsize
% \small
% \hspace{-1.7mm}
\resizebox{\linewidth}{!}{
\begin{tabular}{@{\hskip 0.05in}l@{\hskip 0.05in}|c|c|c|c|c} 
\hline
Method & Filter & top-$k$ & Precision & Recall & F1 \\ 
\hline\hline
\multirow{2}{*}{\emph{BERT-R}} & \xmark & 10 & 39.9 & 50.8 & 44.7 \\
& \cmark & 10 & 46.5 & 45.5 & 46.0 \\
\hdashline
\multirow{4}{*}{\emph{PBERT-R}} & \xmark & 10 & 48.4 & 45.6 & 46.9\\
& \cmark  & 10 & 46.9 & 43.3 & 45.1 \\
& \xmark & 50 &  47.8 & 45.5 & 46.7 \\
& \cmark & 50 & 49.5	& 46.3 &	47.8 \\
\hdashline
\multirow{4}{*}{\emph{SimCSE-R}} & \xmark & 10 & 48.4 & 47.2 & 47.8\\
& \cmark & 10 &  49.4 & 44.8 & 47.0 \\
 & \xmark & 100 & 42.1	& 41.3	& 41.7 \\
& \cmark & 100 &  51.0 & 45.2 & 47.9 \\
\hline
\end{tabular}
}
% \vspace{-2mm}
\caption{Model performances with and without filtering (i.e., w/o the filter model) with top-$k$. In general, without filtering,
augmenting the retrieved candidates enhances recall but may reduce the precision (and hence may not improve the overall F1). Filtering, however, improves the performance, especially with larger top-$k$ candidates. In above, top-100 augmented examples ($\sim$13K total positives after filtering) retrieved by a single retriever perform worse than with top-10 examples by \tool or \toolext (total 7K or 4K positive examples) reported  in Table \ref{table:main-result} in the main paper. 
}
\label{table:filter-impact-appendix}
\end{table}

\clearpage
% \newpage
\section{More Qualitative Examples}
\label{appendix:sec:qual-example}
The below tables show some example retrievals of different models. Retrieved candidates  are distinct from expert annotated ones and can bring auxiliary knowledge to the model.

\begin{table}[!ht]
\centering
% \vspace{4mm}
\resizebox{\linewidth}{!}{%
% \small
\def\arraystretch{1.5}%
\begin{tabular}{p{0.99\linewidth}}
\hline
Q: do you sell my photos to anyone? \\
\hline
% \medskip
\vspace{-6mm}
{\bf Gold:} i) We use third-party service providers to serve ads on our behalf across the Internet and sometimes on the Sites. (ii)
These companies may use your personal information to enhance and personalize your shopping experience with us, to communicate with you about products and events that may be of interest to you and for other promotional purposes. 
 iii) Your use of our Application with that healthcare institution may be subject to that healthcare institution's policies and terms. 
 (iv) We may share personal information within our family of brands. 
(v) From time to time we share the personal information we collect with trusted companies who work with or on behalf of us.
(vi) No personally identifiable information is collected in this process.
(vii) We use third-party service providers to serve ads on our behalf across the Internet and sometimes on our Sites and Apps.
\\
\vspace{-6mm}
% \noskip
{\color{blue}\bf Correct Retrievals:} 
(i) The Application does not collect or transmit any personally identifiable information about you, such as your name, address, phone number or email address.
-(SimCSE-R)
(ii) Some of this information is automatically gathered, and could be considered personally identifiable in certain circumstances, however it will generally always be anonymised prior to being viewed by Not Doppler, and never sold or shared.
-(BERT-R)
(iii) We also use the Google AdWords service to serve ads on our behalf across the Internet and sometimes on this Website.
-(PBERT-R)
(iv) To organ and tissue donation requests: By law, we can disclose health information about you to organ procurement organizations.
-(BERT-R)
\\
\vspace{-6mm}
{\color{red}\bf Incorrect Retrievals:} 
(i) When you upload your photos to our platform or give us permission to access the photos stored on your device, your photo content may also include related image information such as the time and the place your photo was taken and similar “metadata” captured by your image capture device.  -(SimCSE-R)
(ii) These are not linked to any information that is personally identifiable.-(BERT-R)
\\
\hline
\end{tabular}
}
% \vspace{-2mm}
\caption{A fraction of retrieval examples (i).}
\label{table:ret-example-details}
% \vspace{-2mm}
\end{table}

\newpage 
\begin{table}[!ht]
\centering
\vspace{11mm}
\resizebox{\linewidth}{!}{%
% \small
\def\arraystretch{1.5}%
\begin{tabular}{p{0.99\linewidth}}
\hline
Q: who all has access to my medical information? \\
\hline
% \medskip
\vspace{-6mm}
{\bf Gold:} i) Apple HealthKit to health information and to share that information with your healthcare providers. ii) Your use of our Application with that healthcare institution may be subject to that healthcare institution's policies and terms. \\
\vspace{-6mm}
% \noskip
{\color{blue}\bf Correct Retrievals:} 
(i) We may share your information with other health care providers, laboratories, government agencies, insurance companies, organ procurement organizations, or medical examiners.
-(SimCSE-R)
(ii) Do not sell your personal or medical information to anyone.
-(BERT-R)
(iii) Lab, Inc will transmit personal health information to authorized medical providers. 
-(PBERT-R)
(iv) To organ and tissue donation requests: By law, we can disclose health information about you to organ procurement organizations.
-(BERT-R)
\\
\vspace{-6mm}
{\color{red}\bf Incorrect Retrievals:} 
(i) However, we take the protection of your private health information very seriously.  -(SimCSE-R)
(ii) All doctors, and many other healthcare professionals, are included in our database. -(PBERT-R)
(iii) You may be able to access your pet’s health records or other information via the Sites. -(BERT-R)
(iv) will say “yes” unless a law requires us to disclose that health information.-(BERT-R)
(v) do not claim that our products “cure” disease.-(BERT-R)
(vi) Has no access to your database password or any data stored in your local database on your devices.-(BERT-R)
\\
\hline
\end{tabular}
}
% \vspace{-2mm}
\caption{A fraction of retrieval examples (ii).}
\label{table:ret-example-details-ii}
% \vspace{-2mm}
\end{table}

% s: The Application does not collect or transmit any personally identifiable information about you, such as your name, address, phone number or email address.
% Pb: We also use the Google AdWords service to serve ads on our behalf across the Internet and sometimes on this Website.
% b: Some of this information is automatically gathered, and could be considered personally identifiable in certain circumstances, however it will generally always be anonymised prior to being viewed by Not Doppler, and never sold or shared.

% Incorrect:
% Simcse: When you upload your photos to our platform or give us permission to access the photos stored on your device, your photo content may also include related image information such as the time and the place your photo was taken and similar “metadata” captured by your image capture device.
% PrivacyBERT:
% BERT: These are not linked to any information that is personally identifiable. (BERT)

\end{document}